\pdfoutput=1

\documentclass[11pt]{article}

\usepackage{ACL2023}

\usepackage{times}
\usepackage{latexsym}
\usepackage{amsmath}
\usepackage{graphicx}
\usepackage{multirow}
\usepackage{booktabs}
\usepackage{enumitem}
\usepackage[most]{tcolorbox}
\usepackage[T1]{fontenc}

\usepackage[utf8]{inputenc}

\usepackage{microtype}

\usepackage{inconsolata}

\title{\texttt{IntrEx}: A Dataset for Modeling Engagement in Educational Conversations}

\author{Xingwei Tan$^{1 2}$, \bf{Mahathi Parvatham}$^3$, \bf{Chiara Gambi}$^3$, \bf{Gabriele Pergola}$^1$ \\
  $^1$Department of Computer Science, University of Warwick, UK\\
  $^2$School of Computer Science, University of Sheffield, UK\\
  $^3$Department of Psychology, University of Warwick, UK\\
  \texttt{Xingwei.Tan@sheffield.ac.uk}\\
  \texttt{\{Chiara.Gambi, Gabriele.Pergola.1\}@warwick.ac.uk}\\}

\begin{document}
\maketitle
\begin{abstract}
Engagement and motivation are crucial for second-language acquisition, yet maintaining learner interest in educational conversations remains a challenge. While prior research has explored what makes educational \textit{texts} interesting, still little is known about the linguistic features that drive engagement in \textit{conversations}. To address this gap, we introduce \texttt{IntrEx}\footnote{ \url{https://huggingface.co/collections/XingweiT/intrex-68a8f2c97688157066860ae2}}, the first large dataset annotated for interestingness and expected interestingness in teacher-student interactions. Built upon the Teacher-Student Chatroom Corpus (TSCC), \texttt{IntrEx} extends prior work by incorporating sequence-level annotations, allowing for the study of engagement beyond isolated turns to capture how interest evolves over extended dialogues. We employ a rigorous annotation process with over $100$ second-language learners, using a comparison-based rating approach inspired by reinforcement learning from human feedback (RLHF) to improve agreement.
We investigate whether large language models (LLMs) can predict human interestingness judgments. We find that LLMs (7B/8B parameters) fine-tuned on interestingness ratings outperform larger proprietary models like GPT-4o, demonstrating the potential for specialised datasets to model engagement in educational settings. 
Finally, we analyze how linguistic and cognitive factors, such as \textit{concreteness}, \textit{comprehensibility} (\textit{readability}), and \textit{uptake}, influence engagement in educational dialogues.

\end{abstract}

\section{Introduction}

Engagement and motivation are fundamental when learning a second language, influencing both learning outcomes and retention \cite{dornyei2021teaching, masgoret2003attitudes}.
However, while existing studies emphasize the relevance of content \cite{lee2017impact, goris2019effects, tan-etal-2025-cascading}, it is still an open research question how linguistic properties themselves shape engagement in conversations. This is particularly relevant in educational settings, where structured dialogue plays a central role in knowledge transfer, especially as dialogue-based learning environments continue to expand with the use of AI tutors based mainly on dialogue interactions \cite{caines-etal-2022-teacher,doulingomax, tan-etal-2025-safespeech}. 

Existing research has identified linguistic features, such as concreteness, that contribute to the interestingness of texts, often relying on whole-document assessments \cite{sadoski2001resolving,Perg19, pergola-etal-2021-disentangled, Lupo2019AnEO,lee-lee-2023-lftk,nguyen-etal-2024-multi}. Yet, in \textit{educational} conversations, engagement is not static but shaped by thematic continuity, discourse structure, and interaction flow. Whole-document assessments offer limited insight into how interest\footnote{While we use the terms \textit{interest}, \textit{curiosity} and \textit{motivation} interchangeably, we acknowledge nuanced distinctions between these concepts as discussed in previous literature \cite{krapp1994interest,peterson2019curiosity,donnellan2022curiosity}.} develops across multiple exchanges, making it essential to analyze sequence-level discourse features.  

\begin{figure}
    \centering
    \includegraphics[width=0.99\linewidth]{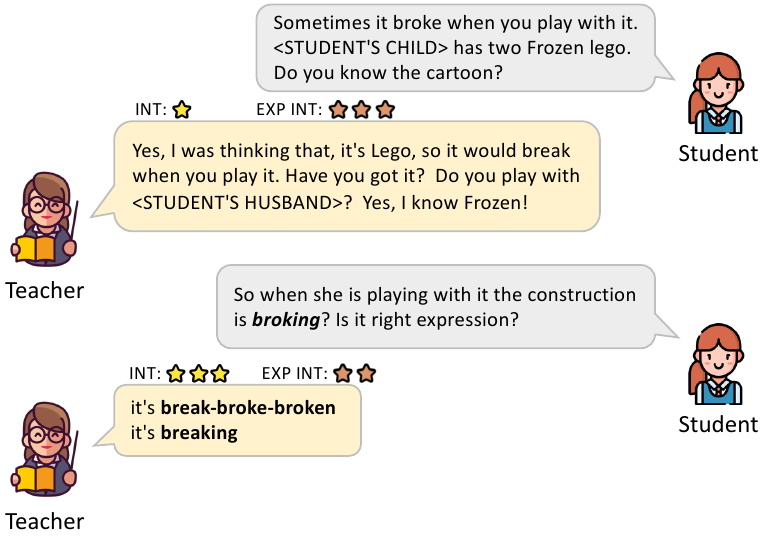}
    \caption{An example of the student-teacher conversations in the IntrEx dataset, along with interestingness (INT) and expected interestingness (EXP INT) scores. \vspace{-12pt}}
    \label{fig:enter-label}
\end{figure}

To investigate this, we build upon the Teacher-Student Chatroom Corpus (TSCC) \cite{caines-etal-2020-teacher, caines-etal-2022-teacher}, the largest dataset of chat-based educational interactions between teachers and second-language learners. The TSCC provides pedagogical discourse annotations, such as \textit{topic opening} and \textit{explanation}. However, while these annotations describe functional aspects of conversation, they do not account for the engagement and interestingness experienced by learners.

To bridge this gap, we introduce \texttt{IntrEx}, a dataset for modelling interestingness and its expectation in educational conversations, resulting from a comprehensive annotation process, involving over 100 second-language learners, to introduce sequence-level annotations of \textit{interestingness} that, for the first time, capture at scale how interestingness and expectations evolve across extended interactions. 
We introduce two key dimensions of conversational engagement: \textit{interestingness}, reflecting the perceived level of engagement, and \textit{expected interestingness}, capturing the anticipated level of engagement. 
These two scores indicate not only what learners found interesting, but also what they expected to find interesting \cite{murayama2022reward}. 
However, assessing interestingness is inherently subjective and prone to variability among annotators. To improve consistency, we thus devised a comparison-based annotation approach inspired by reinforcement learning from human feedback (RLHF) strategies \cite{stiennon2020learning, ouyang2022training}. Instead of rating conversational turns in isolation, annotators compare original utterances with automatically rewritten ‘boring’ versions, leading to higher agreement and reduced noise.

nabled by our new annotations, we conduct two main analyses focused on: (i) the linguistic features that contribute to interestingness, and (ii) the notion of interestingness encoded in LLMs. 
For linguistic features, we analysed \textit{concreteness}, \textit{comprehensibility}, and \textit{uptake} via a wide range of metrics \cite{hosseini-etal-2022-gispy, lyu-pergola-2024-scigispy}, disentangling how each factor influences engagement in teacher-student interactions using linear mixed-effect regressions. 
We conduct experiments on standard vs. instruction-based fine-tuning and analyze how our dataset improves alignment with human interest judgments. Notably, we find that fine-tuned models with only $7$B–$8$B parameters outperform larger models like GPT-4o \cite{openai2024gpt4}, in predicting interestingness in educational dialogues, demonstrating the potential of this new dataset for reward-modelling and for fostering research aimed at improving conversational models for second-language learning.
We then conduct a comprehensive investigation of the ability of LLMs to predict \textit{interestingness}, similarly to human annotators. Recent work (e.g., LaMDA \cite{thoppilan2022}) has incorporated interestingness as an implicit scoring metric, but no existing dataset provides explicit human-labeled annotations for engagement in educational dialogues.

\noindent Our contributions can be summarized as follows:
\begin{itemize}[itemsep=2pt, topsep=1pt]
    \item We introduce the first dataset annotated for interestingness and expected interestingness in educational conversations for second-language learners. We release individual annotator ratings, demographic information, and aggregated scores to ensure dataset transparency. 
    \item We conduct an experimental assessment exploring how our dataset can facilitate the alignment of LLMs with human interest in learning settings. By fine-tuning with IntrEx data, we show that small LLMs have the potential to outperform larger LLMs in predicting conversational engagement.
    \item We conduct an extensive analysis of how linguistic and cognitive factors influence engagement in teacher-student dialogues, specifically focusing on concreteness, comprehensibility (readability), and uptake.
\end{itemize}

\section{Related Work}

\subsection{Conversational Corpora}

Several corpora have been developed to study spoken language in various contexts.
The Cambridge and Nottingham Corpus of Discourse in English (CANCODE) records spontaneous conversations in diverse informal settings and has been used to study the grammar of spoken interaction \cite{carter1997exploring}.
The British National Corpus features transcriptions of spoken conversation captured in settings ranging from parliamentary debates to casual discussion among friends and family \cite{love2017spoken}.
Corpora based on educational interactions, such as lectures and small group discussions, include the widely-used Michigan Corpus of Academic Spoken English (MICASE) \cite{simpson2002michigan}, the TOEFL2000 Spoken and Written Academic Language corpus \cite{biber2004representing}, and the Limerick Belfast Corpus of Academic Spoken English (LIBEL) \cite{o2012applying}.
More specialized corpora include the Why2Atlas Human-Human Typed Tutoring Corpus \cite{rose-etal-2003-comparison}, containing physics tutoring chats, and a corpus of conversations between native speakers and learners of Japanese collected in a VR campus \citet{toyoda2002categorization}.
The Teacher-Student Chatroom Corpus (TSCC) \cite{caines-etal-2020-teacher} and its extended version \cite{caines-etal-2022-teacher} contain chatroom dialogues and are annotated with conversational analysis of sequence types, pedagogical focus, and correction of grammatical errors.
However, none of the aforementioned corpora provide annotations for \textit{interestingness}, which is the focus of our current work.

\subsection{Interestingness}
Researchers across the fields of psychology and education have extensively investigated the factors associated with \textit{interestingness}, here defined following \citet{thoppilan2022}, as the degree to which a text captures attention or sparks curiosity.
\citet{murayama2022reward} discussed the neuro-cognitive mechanisms underlying human interest, suggesting that learning, as an information-seeking behaviour, is driven by reward prediction errors:  Positive prediction errors (rewards exceeding expectations) motivate further exploration, while negative prediction errors diminish motivation. 
\citet{tin2009features}, controlling for various lecture characteristics, found that tangible, personalized, and contextualized content increases learner interest during lectures. 
Other prominent psychological theories suggest a complex relationship between interest and comprehensibility or complexity \cite{wharton1988imagery, dubey2020reconciling, oudeyer2016intrinsic}. 
The ``Goldilocks effect'' \cite{kidd2015psychology} posits that learners prefer stimuli that are neither too simple nor too complex for their current understanding. 
This aligns with the information gap theory \cite{loewenstein1994psychology}, which proposes that the desire for information is fueled by perceived gaps in existing knowledge.  
Stimuli that are too simple present minimal knowledge gaps and thus elicit little interest, while overly complex stimuli create insurmountable gaps, leading to avoidance rather than interest.  
Therefore, comprehensibility likely needs to fall within an optimal range to maximize interest.  
Texts that are either too easy or too difficult will be less engaging than those at an appropriate level of comprehensibility.  
This also implies that highly concrete texts, potentially easier to comprehend, may not correlate positively with interest, as they could be perceived as overly simplistic and thus less engaging.

\begin{figure*}
    [htb]
     \centering
     \includegraphics[width=0.95\linewidth]{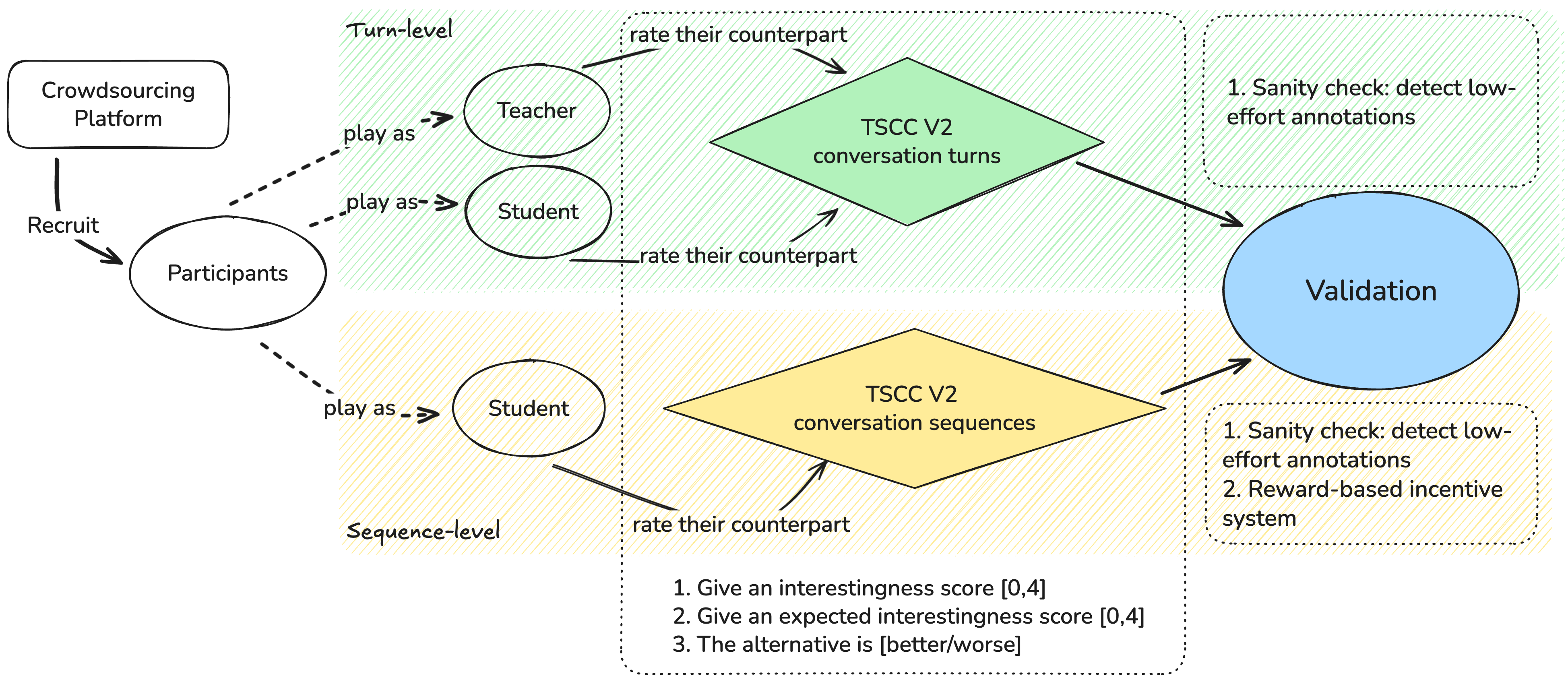}
     \caption{The overall process of the annotation to build \texttt{IntrEx}. \vspace{-12pt}}
     \label{fig:pipeline}
\end{figure*}

\section{The \texttt{IntrEx} Dataset}
\texttt{IntrEx} provides annotations for the conversations in TSCC V2.
This section describes TSCC V2 and the \texttt{IntrEx} annotation process.
Fig. \ref{fig:pipeline} shows the overall process.

\subsection{Background: The TSSC V2 Corpus}
TSCC V2 \cite{caines-etal-2020-teacher, caines-etal-2022-teacher} comprises the conversation histories of one-to-one English learning lessons conducted in online and private chatrooms.
Each lesson typically lasts about an hour.
In total, the dataset contains $260$ conversations involving $2$ teachers and $12$ students.
A \textit{dialogue snippet} is defined as an exchange of messages, generally consisting of one message (a \textit{turn}) from the teacher and one from the student. 
The average number of dialogue snippets per conversation is $52$ (ranging from $26$ to $90$).
The conversations are manually annotated for sequence type, indicating major or minor shifts in the conversational flow (e.g., \textit{opening}, \textit{exercise}, \textit{redirection}), and teaching focus, which identifies the skills targeted within each sequence.

\subsection{Task Definition}
\label{sec:task_definition}
We annotate two types of engagement scores: \textit{interestingness} and \textit{expected interestingness}.
They are both integer values ranging from $0$ to $4$.
Following \citet{thoppilan2022}, we define \textit{Interestingness} as the degree to which the current message captures the annotator's attention or sparks curiosity.
\textit{Expected interestingness} represents the annotator's prediction of their \textit{interestingness} level after receiving the next message.
Since annotators provide this score before seeing the next message, it reflects speculative engagement expectations rather than actual interest.
By collecting \textit{expected interestingness}, we can quantify how new topic introductions or discourse shifts exceed, meet, or fall short of expectations. This allows us to analyze the role of anticipation in shaping the potential engagement for the students, offering insights into how conversational structure influences perceived interestingness.
The definitions are presented to the annotators as shown in Figure \ref{fig:prolific_sequence} and \ref{fig:prolific_turn} in Appendix \ref{sec:appendix_UI}.
Presenting explicit yet intuitive definitions helped reduce potential bias from overly prescriptive instruction, while still allowing for natural inter-individual variability in how engagement is perceived. To account for this subjectivity, each part of the dataset was independently rated by three different annotators, ensuring a diverse range of judgments.

There are two types of annotations in \texttt{IntrEx}: turn-level and sequence-level.

\noindent \textbf{Turn-level Annotations.} During the turn-level annotation task, half of the annotators imagined themselves as the teacher in the conversation and rated the interestingness of the student’s responses. The other half imagined themselves as the student in the conversation and rated the interestingness of the teacher’s responses.
Our annotation platform (App. \ref{sec:appendix_UI}, Fig. \ref{fig:ui_screenshot}) displays the conversation page by page, where each page contains a dialogue snippet: a teacher turn and a student turn.
Annotators were asked to assign an interestingness score and an expected interestingness score based on the turn from the other party. 
However, we observed that interestingness scores often remained unchanged across consecutive messages, particularly within the same topic. This led to annotator fatigue, reducing annotation quality and lowering inter-annotator agreement. Additionally, engagement in conversations is not always localized to a single turn; it evolves over longer sequences, making turn-level scores too fine-grained to capture conversational dynamics.

\noindent \textbf{Sequence-level Annotations.} To mitigate this, we segmented each conversation into sequences based on the labels provided by the TSCC V2 dataset.
The sequence type labels indicate major conversational shifts, such as a change in teaching contents or in discourse structure. 
A new sequence is initiated whenever a message's associated sequence type differs from that of the preceding message.  
This approach ensures that each sequence represents distinct segments of the lesson, such as \textit{homework}, \textit{clarification}, or \textit{closing}.
Instead of evaluating each turn, annotators provide an \textit{interestingness} and an \textit{expected interestingness} score for a whole sequence, substantially reducing their workload. 
In this setting, the annotators are only required to rate teachers' messages in the sequence-level annotation, because (i) teachers deliver significantly more content, (ii) and the sequence type labels are often about teaching content. 

\begin{table*}[htb]
  \begin{center}
  {\small\begin{tabular}{lcccccccc}
  \toprule
  \multirow{2}{*}{\bf Level} & \multicolumn{3}{c}{\bf Count}  & \multicolumn{2}{c}{\bf Average Score} & \multicolumn{3}{c}{\bf AC2 Agreement}\\
\cmidrule(lr){2-4} \cmidrule(lr){5-6} \cmidrule(lr){7-9}
   & \bf Row & \bf Conv & \bf Annotator& \bf INT & \bf EXP INT & \bf INT & \bf EXP INT & \bf INT \& EXP INT\\ 
  \midrule 
 Turn       & $7,118$ & $65$ & $96$ & $2.10_{\textcolor{gray}{0.77}}$ & $2.00_{\textcolor{gray}{0.71}}$ & $0.40_{\textcolor{gray}{0.13}}$ & $0.39_{\textcolor{gray}{0.15}}$   & $0.39_{\textcolor{gray}{0.13}}$\\
 Sequence   & $5,801$ & $259$ & $48$& $1.35_{\textcolor{gray}{0.80}}$ & $1.42_{\textcolor{gray}{0.74}}$ & $0.58_{\textcolor{gray}{0.14}}$ & $0.52_{\textcolor{gray}{0.15}}$   & $0.55_{\textcolor{gray}{0.13}}$\\

  \bottomrule
  \end{tabular}}
  \end{center}
  \caption{The number of turns/sequences, conversations, and annotators. The mean and standard deviation of the average interestingess (INT) and expected interestingness (EXP INT) acoress the $3$ annotators that rate each conversation. The average AC2 agreement of the ratings for the turn/sequence-level annotations of INT and EXP INT. The gray values are the standard deviations.}
  \label{table:aggreement_levels}
\end{table*}

\subsection{Data Collection and Validation}

\paragraph{Annotation Design and Recruitment Strategy}
Participants were recruited via Prolific\footnote{\href{http://www.prolific.com}{prolific.com}}.
Prolific enables us to screen potential participants based on their self-reported language backgrounds.  
To avoid bias, we restricted participation to individuals who learned English as a second language and were not raised in an English-speaking environment, because second language learning is vastly different from first language acquisition. 
Native speakers may never have experienced conversations like those in the TSCC corpus, and therefore cannot fully understand why a learning point is interesting to a second language learner. For example, if a teacher uses simple terms tailored to the student’s proficiency level, a native speaker annotator might score it as very low in interestingness because they would judge it as too easy from their own perspective.

Upon registering for the study, all participants were provided with detailed guidelines and informed that the data would be anonymized and made available to researchers.
Participants were then directed to our locally hosted annotation platform, which was built using the open-source software \textit{doccano}\footnote{\href{https://github.com/doccano/doccano}{https://github.com/doccano/doccano}}.
Participants were instructed to assume the role of either teacher or student within a given conversation, and to base their ratings on that role, rather than their personal preferences. No participant was assigned both roles within the same conversation.  

\noindent \textbf{Annotator Compensation} A total of $96$ participants completed turn-level annotations, while another $48$ participants annotated at the sequence-level.  
Turn-level annotators were compensated at a rate of $7$£ per hour. Sequence-level annotators received $8$£ per hour, with an additional $3$£ bonus for high alignment within their annotation group.

\paragraph{Enhancing Annotation Quality}

We initially conducted a pilot study at the turn-level. As is common in subjective annotation tasks involving engagement or affective judgments, we observed relatively low inter-annotator agreement and high annotator effort, particularly when rating individual messages in isolation. These findings are consistent with prior work highlighting the challenges of obtaining consistent judgments in tasks involving nuanced, interpretive constructs such as interest or curiosity \cite{rodriguez-etal-2020-information, rottger-etal-2022-two}. To increase reliability and mitigate annotator fatigue, we implemented two strategies aligned with the literature in preference learning \cite{brown2020languagemodelsfewshotlearners}.

First, each annotation task included two reference conversations for calibration, selected from examples where previous annotators had shown high agreement. These \textit{reference conversations} served as calibration examples, covering a diverse range of engagement levels and linguistic features (e.g., topic shifts, complexity). 
The first $11$ turns from each reference conversation were prepended and appended to each annotation task.

Second, to reduce subjectivity in scoring, we adopted a comparison-based approach inspired by Reinforcement Learning from Human Feedback (RLHF) \cite{stiennon2020learning, ouyang2022training}. Annotators rated each message alongside a “boring” alternative generated by GPT-4o, an automatically rewritten version that preserved content while removing engaging linguistic features (see the example below and in Appendix \ref{appendix:boring}). The prompts are shown in Appendix \ref{appendix:prompts}. This setup allowed annotators to judge interestingness in relative rather than absolute terms, a method shown to reduce cognitive load and improve consistency in preference-based tasks \cite{clark2018rate}.

    \begin{tcolorbox}[colback=yellow!10, colframe=orange!60!black, title=Original Messages]
    \small it's break-broke-broken

    it's breaking
    \end{tcolorbox}
    
    \begin{tcolorbox}[colback=green!5, colframe=green!60!black, title=Boring Alternative]
    \small The correct verb forms for "break" are: break, broke, broken.

    it is currently in the process of breaking
    \end{tcolorbox}

\paragraph{Sequence-Level Annotations}
Despite the aforementioned measures, our pilot studies confirmed that turn-level annotation remained more variable and cognitively demanding than sequence-level annotation. Given that engagement in dialogue often builds across multiple turns, we therefore prioritized sequence-level engagement annotations as they provide a more reliable and informative representation of conversational engagement focused on multi-turn discourse modelling.

To conduct this annotation, we leveraged TSCC V2 labels to segment conversations based on teaching content, ensuring that each sequence reflected a cohesive unit of interaction. As described in Section \ref{sec:task_definition}, annotators assigned interestingness and expected interestingness scores to each sequence. To enhance annotation consistency and agreement, also in this case we also incorporated a comparison-based alternative, where annotators evaluated original turns alongside GPT-3.5-generated rewrites to facilitate more consistent engagement judgments.

\begin{table*}[htb]
  \begin{center}
  {\small\begin{tabular}{lccccr}
  \toprule
   \bf Level & \bf CEFR & \bf Count & \bf Average Age & \bf Gender  & \bf First Language (Top 3)\\ 
  \midrule 
 \multirow{4}{*}{Turn} & \textit{C2} & $13$  & $28.08$ & M ($7$), F ($6$)   & Greek ($4$), Polish ($3$), Portuguese ($1$)\\
 &\textit{C1} & $36$  & $25.78$ & M ($25$), F ($11$) & Polish ($14$), Portuguese ($8$), Italian ($5$)\\
 &\textit{B2} & $23$  & $30.13$ & M ($10$), F ($13$) & Polish ($14$), Italian ($6$), Portuguese ($1$)\\
 &\textit{A2} & $3$   & $33.33$ & M ($1$), F ($2$)   & Polish ($1$), Greek ($1$), Spanish ($1$)\\
 \midrule
 \multirow{4}{*}{Sequence} & \textit{C2} & $9$ & $29.56$ & M ($6$), F ($3$)   & Portuguese ($5$), Greek ($2$), Polish ($1$)\\
 &\textit{C1} & $24$& $27.13$  & M ($12$), F ($12$)& Portuguese ($8$), Polish ($4$), Italian ($4$)\\
 &\textit{B2} & $14$& $28.21$  & M ($6$), F ($8$)  & Polish ($7$), Estonian ($2$), Portuguese ($2$)\\
 &\textit{A1} & $1$ & $24.00$  & M ($1$)           & Polish ($1$)\\
  \bottomrule
  \end{tabular}}
  \end{center}
  \caption{The average age and the distributions of gender and first language grouped by the CEFR level of the annotators in the turn-level and sequence-level annotation.}
  \label{table:proficiency_turn_level}
\end{table*}

\paragraph{Quality Control and Agreement Measures}
To maintain annotation quality, we implemented sanity checks and a reward-based incentive system.

First, we conducted a sanity check to identify low-effort annotations by detecting sequences where annotators assigned identical scores to $10$ or more consecutive turns were excluded from the dataset.
Second, each message was annotated by three annotators, and agreement was measured using Gwet’s AC2 metric \cite{Gwet2008ComputingIR,Gwet2014HandbookOI}, chosen for its stability in cases of variable annotator agreement. AC2 is computed as:

\begin{equation}
    \text{AC2} = 1 - \frac{\sum w_{ij} \cdot f_{ij}}{\sum w_{ij} \cdot e_{ij}}
\end{equation}

\noindent where $f_{ij}$ is the frequency with which the first rater gives score $i$ and the second rater gives score $j$, $e_{ij}$ is the expected frequency under the assumption of statistical independence, and $w_{ij}$ is the weight for agreement between scores $i$ and $j$.
Using an identity matrix as $W$ would treat both small ($3\rightarrow4$) and large ($1\rightarrow4$) differences equally as complete disagreement. 
Thus, we instead applied a linear weight matrix (Eq.  \ref{eq:gwet_matrix}), which makes disagreement decrease linearly with the distance between categories.
It is also necessary to clarify that this choice of weight matrix is part of the evaluation metric and is independent from the data.
\vspace{-8pt}

\begin{equation}
W_{\text{AC2}} =
{\small \begin{bmatrix}
1.00 & 0.75 & 0.5 & 0.25 & 0.00 \\
0.75 & 1.00 & 0.75 & 0.50 & 0.25 \\
0.50 & 0.75 & 1.00 & 0.75 & 0.50 \\
0.25 & 0.50 & 0.75 & 1.00 & 0.75 \\
0.00 & 0.25 & 0.50 & 0.75 & 1.00 \\
\end{bmatrix}}
\label{eq:gwet_matrix}
\end{equation}

Finally, a reward system was implemented at the sequence-level stage.  
Annotator groups (of three) achieving an AC2 score of $0.5$ or higher on their \textit{interestingness} annotations received a £$3$ bonus per annotator and were prioritized for future tasks.

Table \ref{table:aggreement_levels} reports the dataset statistics, average scores, and AC2 agreement of the turn-level and sequence-level annotations.
We compute the AC2 of $3$ annotators jointly in each annotation task and then average them across all the tasks.
Sequence-level annotations have a significantly higher level of agreement than turn-level on INT ($p=3.19e^{-8}$), EXP INT ($p=6.82e^{-5}$), and combined scores ($p=5.12e^{-7}$), indicating the difficulty in evaluating interestingness on a fine-grained level.

\subsection{Dataset Statistics and Demographics}

Turn-level annotations were collected for $64$ conversations ($25\%$ of the corpus), while sequence-level annotations, focusing solely on teacher messages, were collected for $259$ conversations \footnote{Annotator information can be downloaded from \url{https://github.com/Xingwei-Tan/IntrEx}}. 
A total of $5,801$ sequences were annotated, with an average of $22.4$ sequences per conversation (ranging from $8$ to $41$).
Turn-level annotations consist of $7,118$ pairs of scores in total.

Table \ref{table:proficiency_turn_level} presents the self-reported age, gender, first language, and English proficiency (measured using the Common European Framework of Reference, CEFR) of participants in the turn- and sequence-level annotation tasks.
In both cases, most annotators identified as B2 or C1 level, reflecting an upper-intermediate to advanced command of English.
This distribution was intentional, as it aligns with the proficiency levels of students in the original TSCC dataset (i.e., mostly from B1 to C2), ensuring that annotators could reliably adopt the perspective of the original students in the conversations.
We also found that conversations with learners of higher proficiency were rated as more interesting, and also that ratings were higher when the annotator's proficiency matched the proficiency of the learner in the original conversation. 
Nonetheless, it is possible that the skew towards annotators with relatively high proficiency may introduce some bias, as annotators not actively engaged in language learning might have perceived the teacher’s instruction as overly simplistic. Additionally, the recruitment platform, Prolific, which is primarily used in Europe and North America, naturally favors users with moderate to advanced English proficiency.

\section{Experiments}
In this section, we leverage the new \texttt{IntrEx} dataset to conduct an experimental assessment of engagement in educational dialogue, focusing on two complementary goals: (i) evaluating the extent to which LLMs align with human judgments of interestingness, and (ii) identifying the linguistic features that contribute to perceived engagement.

\begin{table}
\centering
{\small \begin{tabular}[htb]{lr}
\toprule
\bf Predictor & \bf AC2 \\
\midrule
Random (Gaussian) & $0.3491$\\
\midrule
GPT-4 & $0.4421$\\
GPT-4o & $0.4657$\\
\midrule
Mistral-7B-Base & $0.2129$\\
Llama3-8B-Base & $0.2584$\\
Mistral-7B-Instruct & $0.3608$\\
Llama3-8B-Instruct & $0.3646$\\
Mixtral-8$\times$7B-Instruct & $0.4549$\\
Mistral-7B-Base (\texttt{-IntrEx}) & $0.1587$\\
Llama3-8B-Base (\texttt{-IntrEx}) & $0.2340$\\
Mistral-7B-Instruct (\texttt{-IntrEx}) & $\textbf{0.5142}$\\
Llama3-8B-Instruct (\texttt{-IntrEx}) & $0.5139$\\
\bottomrule
\end{tabular}}
\caption{The average AC2 agreement between the LLMs and the human rating of interestingness. The last four rows are models fine-tuned on \texttt{IntrEx}.}
\label{tab:llm_prediction}
\end{table}

\begin{table*}[htb!]
\centering
{\fontsize{9.5pt}{10pt}\selectfont \begin{tabular}[t]{llrrrrrr}
\toprule
\multirow{2}{*}{\bf Level} & \multirow{2}{*}{\bf Feature} & \multicolumn{3}{c}{\bf Interestingness}  & \multicolumn{3}{c}{\bf Expected Interestingness}\\
\cmidrule(lr){3-5} \cmidrule(lr){6-8}
&  & \multicolumn{1}{c}{\bf B} & \multicolumn{1}{c}{\bf SE} & \multicolumn{1}{c}{$t$} & \multicolumn{1}{c}{\bf B} & \multicolumn{1}{c}{\bf SE} & \multicolumn{1}{c}{$t$} \\
\midrule
\multirow{13}{*}{Turn}&(Intercept) & $2.085$ & $0.068$ & $30.718$ & $1.977$ & $0.067$ & $29.733$\\
&\textbf{Concreteness} & $-0.039$ & $0.009$ & $\bf{-4.478}$ & $-0.018$ & $0.009$ & $\bf{-2.060}$\\
&\textbf{Coleman-Liau index} & $0.029$ & $0.011$ & $\bf{2.618}$ & $0.030$ & $0.011$ & $\bf{2.776}$\\
&\textbf{Smog index} & $0.030$ & $0.010$ & $\bf{3.035}$& $0.022$ & $0.010$ & $\bf{2.316}$\\
&Automated readability index & $0.004$ & $0.019$ & $0.201$ & $-0.007$ & $0.019$ & $-0.364$\\
&Spache readability & $-0.010$ & $0.016$ & $-0.594$ & $-0.011$ & $0.016$ & $-0.713$\\
&\textbf{GIS} & $0.020$ & $0.009$ & $\bf{2.384}$ & $0.009$ & $0.008$ & $1.115$\\
&\textbf{GIS}$\textsuperscript{2}$ & $-0.028$ & $0.007$ & $\bf{-3.996}$ & $-0.013$ & $0.007$ & $-1.808$\\
&\textbf{Lexicon count} & $0.197$ & $0.010$ & $\bf{19.036}$ & $0.177$ & $0.010$ & $\bf{17.315}$\\
&\textbf{Lexicon count}$\textsuperscript{2}$ & $-0.117$ & $0.008$ & $\bf{-15.024}$ & $-0.093$ & $0.008$ & $\bf{-12.051}$\\
&\textbf{LCS} & $0.028$ & $0.008$ & $\bf{3.615}$ & $0.009$ & $0.008$ & $1.163$\\
&\textbf{Student-uptake-teacher} & $0.020$ & $0.007$ & $\bf{2.807}$ & $0.015$ & $0.007$ & $\bf{2.080}$\\
&Cosine similarity & $-0.015$ & $0.008$ & $-1.933$ & $-0.013$ & $0.008$ & $-1.684$\\
\midrule
\multirow{13}{*}{Sequence}&(Intercept) & $1.379$ & $0.050$ & $27.818$                     & $1.425$ & $0.058$ & $24.586$\\
&Concreteness & $0.000$ & $0.012$ & $-0.039$                  & $0.018$ & $0.011$ & $1.597$\\
&\textbf{Flesch reading ease} & $0.459$ & $0.039$ & $\bf{11.891}$                   & $0.319$ & $0.037$ & $\bf{8.694}$\\
&\textbf{Flesch-Kincaid grade} & $0.677$ & $0.066$ & $\bf{10.229}$                   & $0.539$ & $0.063$ & $\bf{8.566}$\\
&\textbf{Coleman-Liau index} & $0.238$ & $0.025$ & $\bf{9.528}$                   & $0.201$ & $0.024$ & $\bf{8.453}$\\
&Dale-Chall readability & $0.009$ & $0.016$ & $0.566$                 & $0.025$ & $0.015$ & $1.691$\\
&Gunning fog & $-0.026$ & $0.029$ & $-0.914$                  & $-0.017$ & $0.027$ & $-0.612$\\
&\textbf{Smog index} & $0.276$ & $0.010$ & $\bf{27.897}$                    &$0.194$ & $0.009$ & $\bf{20.617}$\\
&\textbf{Automated readability index} & $-0.195$ & $0.039$ & $\bf{-5.038}$                  & $-0.254$ & $0.037$ & $\bf{-6.907}$\\
&\textbf{Spache readability} & $-0.093$ & $0.036$ & $\bf{-2.622}$                 & $-0.036$ & $0.034$ & $-1.073$\\
&\textbf{Student-uptake-teacher} & $-0.055$ & $0.009$ & $\bf{-6.254}$                    & $-0.035$ & $0.008$ & $\bf{-4.186}$\\
&\textbf{propTinS} & $0.113$ & $0.010$ & $\bf{11.291}$                  & $0.075$ & $0.009$ & $\bf{7.934}$\\
&\textbf{Cosine similarity} & $-0.046$ & $0.011$ & $\bf{-4.219}$                   & $-0.088$ & $0.010$ & $\bf{-8.554}$\\
\bottomrule
\end{tabular}}
\caption{Combined model predicting turn-level or sequence-level ratings respectively. All predictors are scaled, so coefficients (\textbf{B}) are standardised. \textbf{SE} is the standard error of the coefficient. $t$ is the t-statistic. Significant predictors are highlighted in bold (i.e., $|t|\ge2$).} \vspace{-8pt}
\label{tab:combined_turn_expint}
\end{table*}

\subsection{Reward Modeling with \texttt{IntrEx}}
First, we investigate how LLMs, with or without instruction-tuning, assess conversational interestingness.
We then fine-tuned Llama3-8B \cite{dubey2024llama} and Mistral-7B \cite{jiang2023mistral7b} to predict human \textit{interestingness} scores using \texttt{IntrEx} as a multi-class classification task. 
The model inputs consisted of the conversation history preceding the target message to be scored. The target message was delimited by \texttt{<target-of-rating>} and \texttt{</target-of-rating>} tags.
The average of the three annotators' scores for each message, rounded to the nearest integer, served as the ground truth label.  Sequence-level annotations comprised the training set, while turn-level annotations formed the test set.

Table \ref{tab:llm_prediction} compares the performance of our fine-tuned models against the off-the-shelf LLMs using in-context prompting (prompts are reported in Appendix \ref{appendix:prompts}).  
As an additional baseline, we include a random number generator which samples from $N(2,1)$ and rounds to the nearest integer.
The random baseline has an AC2 agreement of $0.3491$ with respect to humans.
The frontier proprietary LLMs GPT-4 and GPT-4o achieved AC2 scores of $0.4421$ and $0.4657$, respectively, outperforming the smaller Mistral-7B and Llama3-8B.  
However, after fine-tuning on the \texttt{IntrEx}, Llama3-8B-Instruct and Mistral-7B-Instruct surpass the much larger GPT-4, GPT-4o, and Mixtral \cite{jiang2024mixtralexperts}. 
This result also shows that the fine-tuned models demonstrate strong generalization capabilities, generating fine-grained, turn-level feedback despite being trained on coarser, sequence-level data.  

Notably, base LLMs performed worse than the random baseline, even after fine-tuning (Table \ref{tab:llm_prediction}). This suggests that - since they lack pre-training exposure to explicit rating tasks - base models struggle with instruction following in rating-based contexts. In contrast, instruction-tuned models (e.g., Llama3-8B-Instruct) showed greater alignment with human interestingness scores, emphasizing the importance of instruction tuning for engagement modeling.

\subsection{Linguistic Predictors of Human Interest}

In our second analysis, we examine how linguistic factors, such as concreteness, comprehensibility, and uptake, contribute to perceived engagement in teacher–student interactions.
To examine the individual and combined effects of concreteness, comprehensibility, and uptake on interestingness, we computed a range of metrics (Table \ref{tab:combined_turn_expint}). 
\begin{itemize}[noitemsep, topsep=2pt]
    \item \textbf{Concreteness} was measured as the average concreteness rating of the content words in each turn or sequence. The concreteness ratings are from the MRC Psycholinguistic Database \cite{Wilson1988MRC}. We selected the metric based on the MRC database due to its broader lexical coverage across the conversational vocabulary in our dataset.
    \item \textbf{Gist Inference Score (GIS)} is based on the principle that more comprehensible texts facilitate gist extraction: the ability to form an abstract representation of the content without needing to retain \textit{verbatim} details. Because the GIS score formula inherently penalizes concreteness, and given their moderate negative correlation, we analyzed concreteness separately as well as in conjunction with the GIS score.  Critically, the GIS score has been empirically validated, at least for medical texts, with evidence suggesting that higher GIS scores correlate with greater human comprehension \cite{wolfe2019theoretically, dandignac2020gist}.
    \item \textbf{Flesch reading ease} indicates how easy the text is to read. The lower the score, the tougher it is to read. The formula takes into account the total number of words, sentences, and syllables.
    \item \textbf{Flesch-Kincaid grade} is similar to the Flesch Reading Ease, using the same variables in the formula, but it is more extensively used in education. A higher score indicates the text is suitable for a higher grade, thus more difficult.
    \item \textbf{Coleman-Liau index} depends on the average number of letters per 100 words and the average number of sentences per 100 words.
    \item \textbf{Dale-Chall readability} uses a list of 3000 words that groups of fourth-grade American students rated on a difficulty scale.
    \item \textbf{Gunning fog index} determines the years of formal education required by the reader to understand the text. It is based on the number of words, sentences and complex words.
    \item \textbf{Smog index} uses the number of sentences and polysyllables to determine the years of education needed to understand text.
    \item \textbf{Automated readability index} determines the comprehensibility of a text based on the number of characters, words and sentences.
    \item \textbf{Spache readability} rates the comprehensibility of text based on the number of common everyday words it contains.
    \item \textbf{Longest Common Subsequence (LCS)} identifies common words in consecutive student and teacher turn.
    \item \textbf{Student-uptake-teacher} assess the extent to which the next speaker built upon the previous speaker’s contribution for two consecutive turns. This metric was derived from a BERT model \cite{wang-demszky-2024-edu} trained on a dataset of educational conversations \cite{demszky2023improving}. The score represents the student's uptake of the teacher's turn.
    \item \textbf{propTinS} is the proportion of words the teacher repeated from the student's message.
    \item \textbf{Cosine similarity} is the similarity between \textit{mxbai-embed-large-v1} \citep{emb2024mxbai} embeddings of consecutive teacher/student turns.
\end{itemize}

\noindent \textbf{Concreteness.} We found that concreteness was negatively correlated with interestingness, which is in line with emerging evidence that overly simple content may reduce engagement \cite{oudeyer2016intrinsic, dubey2020reconciling}. This finding is consistent across turn- and sequence-level.

\noindent \textbf{Comprehensibility.} Comprehensibility was assessed via a wide-range of metrics (i.e., GIS, Flesch reading ease, Flesch-Kincaid grade, Coleman-Liau index, Dale-Chall readability, Gunning fog index, Smog index, Automated readability index, and Spache Readability)
 We first run comparisons between a simple linear regression model and a regression model including both a linear and a quadratic predictor (linear and quadratic predictors were always orthogonal polynomials) for each comprehensibility feature separately. We then included all significant (i.e., $|t|\ge2$) predictors in a single linear mixed-effect model with random (intercept) effects for conversation and annotation project (i.e., subset of $3$ annotators who completed the same annotation task). This was done twice: with either interestingness or expected interestingness as the outcome variable, and we only retained those metrics that were robust predictors in both models. 

At both turn- and sequence-level, several readability metrics were positively related to interest, suggesting that conversations that are more suitable for a higher grade tended to be more interesting for our annotators. Note that, at the sequence-level, some readability metrics had instead a negative relation with interest, suggesting the relation between comprehensibility and interest is complex. In addition, but only at the turn-level, we observed an inverted U-shaped relation with turn length, i.e., lexicon count, the GIS score, and gunning fog, a traditional readability metric. Such an inverted U-shaped relation is predicted by several computational and psychological theories of interest \cite{oudeyer2016intrinsic,dubey2020reconciling}, and suggests that both very simple and very complex messages lower interest levels, compared to messages that challenge the reader ``just enough''. 

\noindent \textbf{Uptake.} While concreteness and comprehensibility are widely applicable to different types of text, the third feature we investigated, \textit{uptake}, is specific to conversational turns. Qualitative research using conversation analysis has long emphasized the importance of uptake, particularly in educational settings \cite{huth2011conversation, walsh2013classroom}. 
The degree to which a teacher reuses elements of a student's turn provides reassurance that the student is being heard and encourages further interaction, thereby enhancing the learning process and boosting student confidence. 
\citet{see2019makes} explored whether controlling for response-relatedness (measured via embedding similarity) improves human judgments of \textit{interestingness} in human-LSTM network conversations. 
Inspired by this work, we also investigated response-relatedness, calculating several uptake measures: LCS, propTinS, model-based estimates of conversational uptake, and embedding-based cosine similarity between turns. 

In both turn-level and sequence-level analyses, we found positive correlations between some uptake measures and interest (\textit{LCS} and \textit{student-uptake-teacher} in turn-level analyses, \textit{propTinS} in sequence-level analyses).
However, cosine similarity was a negative predictor in both turn-level and sequence-level analyses, and student-uptake-teacher was a negative predictor in sequence-level analyses. This seemingly contradictory finding could suggest that introducing novel ideas, which lowers similarity between successive turns or sequences, may also be important to promote interest, and that uptake of the teacher by the student may index a different phenomenon when it is computed at the turn-level vs. the sequence-level.

\section{Conclusion}

We introduced \texttt{IntrEx}, the first dataset that annotates both interestingness and expected interestingness for teacher–student dialogues, extending TSCC with sequence-level labels to track how engagement evolves across a lesson. We collected ratings from more than a hundred L2 learners and designed a comparison-based annotation, achieving substantially higher reliability at the sequence-level than the turn-level. Small, instruction-tuned LLMs fine-tuned on \texttt{IntrEx} surpassed GPT-4/4o in predicting human interest ratings, demonstrating the relevance and impact of hiqh-quality labels for the task. Future work will be able to leverage \texttt{IntrEx} for training reward models for generation, and extend the approach to new learner groups and domains.

\section*{Limitations}
\paragraph{Annotator Proficiency Bias}
While our annotators were second-language English speakers, most had at least B2 proficiency. This choice was intentional to align with the proficiency level of the original students in the TSCC dataset and to ensure annotators could reliably interpret the conversations and follow the task instructions. However, this may underrepresent the perspective of lower-proficiency learners, whose criteria for engagement might differ. Capturing those perspectives would require task simplification and dedicated data collection, which we consider an important direction for future work.

\paragraph{Subjectivity of Interestingness}
Interestingness is inherently subjective, and inter-annotator agreement is limited by individual variation in interests and background knowledge. We mitigated this through (i) targeted recruitment of language learners with sufficient comprehension skills, (ii) consistent task instructions emphasizing pedagogical engagement rather than personal preference, and (iii) a comparison-based annotation framework that anchors judgments against low-interest alternatives. Nonetheless, subjectivity cannot, and probably should not, be eliminated entirely, and our results should be interpreted accordingly.

\paragraph{Generalizability to Other Domains}
The \texttt{IntrEx} dataset focuses exclusively on teacher–student interactions within English-as-a-second-language learning contexts. As such, the findings may not fully generalize to informal peer conversations, other educational domains (e.g., math), or multilingual scenarios. While our controlled setting allowed us to isolate linguistic drivers of engagement, future work could extend this approach to broader conversational contexts, including long-term dialogue or cross-domain applications.

\paragraph{Model Evaluation Scope}
Our evaluation of LLMs is limited to predicting interest ratings on human-authored conversations. While this serves as a proxy for alignment with human judgments, we do not assess whether fine-tuned models can generate more engaging conversations themselves. Exploring generation quality and incorporating direct preference optimization (e.g., via DPO) would require new rounds of human evaluation and fall outside the scope of this initial dataset release.

\section*{Ethics Statement}
Prior to recruiting participants, we obtained ethics approval from our institutional review board, with the understanding that all annotations would be anonymized before public release.  While demographic data were collected, these data are not linked to individual participants.  A clear withdrawal procedure was provided to all participants.  Annotators were compensated fairly for their work, commensurate with the task's complexity.

\section*{Acknowledgements}
Xingwei was supported by the Warwick Chancellor's International Scholarship while conducting this work. Chiara Gambi was supported by a Leverhulme Trust Research Project grant (RPG-2023-067).
Gabriele Pergola was partially supported by the ESRC-funded project \textit{Digitising Identity: Navigating the Digital Immigration System and Migrant Experiences}, as part of the Digital Good Network. 
This work was conducted on the Sulis Tier-2 HPC platform hosted by the Scientific Computing Research Technology Platform at the University of Warwick. Sulis is funded by EPSRC Grant EP/T022108/1 and the HPC Midlands+ consortium.

We thank the anonymous reviewers for their helpful comments. The writing of this paper received minor language-polishing suggestions from Gemini.
In addition, parts of our experimental code were drafted or refactored with assistance from GitHub Copilot; all final implementations were manually reviewed and verified by the authors.

\bibliography{anthology,custom}
\bibliographystyle{acl_natbib}

\newpage
\clearpage

\appendix

\section{Fine-tuning LLMs on Human Interest Annotations}
We tested several settings of training models to predict the interestingness scores.
Our first attempt is to train the models to predict the scores as a regression task, but the models often ignore the range of the scores, producing scores lower than $0$ or greater than $4$.
We also tried adding instructions to the input to describe that ``Please rate the level of interestingness of the given message based on the context of the whole conversation.'', but the resulting models are inferior than the ones trained without this instruction.
We compared the performance of training with the base model versus the instruction-tuned version of Llama3 and Mistral, and found the instruction-tuned models have higher agreement.
The Mistral and Llama3 were trained for $3$ epochs with a learning rate of $5e-6$.

\section{Boring Alternative}
\label{appendix:boring}

    \begin{tcolorbox}[colback=yellow!10, colframe=orange!60!black, title=Origin Conversation]
    \small TEACHER: OK good...this is what means to be addicted i.e. you can't stop ...good! OK so a bit more about how/why this is bad.....!

STUDENT: as a consequence, young people don't have time to do some useful things for their life, such as being social and reading.

TEACHER: OK great .... a clear reason about why this is bad ...hold on a sec....
    \end{tcolorbox}
    
    \begin{tcolorbox}[colback=green!5, colframe=green!60!black, title=Boring Alternative]
    \small TEACHER: Please elaborate further on the negative implications of this behavior.

STUDENT: as a consequence, young people don't have time to do some useful things for their life, such as being social and reading.

TEACHER: Please wait momentarily while I process the information...
    \end{tcolorbox}

\section{Prompts}
\label{appendix:prompts}

\begin{tcolorbox}[colback=gray!10, colframe=black!50, title=Prompt for generating boring alternatives]
    \small Instruction: given a text message from a teaching session between a teacher and a student, please provide a more straightforward and less engaging version. Strip away any colourful language or additional context to make the message as boring as possible. Please keep the main information from the message.
\end{tcolorbox}

The prompt used for instructing off-the-shelf instruction-tuned LLMs to rate the turn-level INT and EXP INT:
\begin{tcolorbox}[colback=gray!10, colframe=black!50, title=Prompt for instructing instruction LLMs]
    \small Can I please ask you to rate a student-teacher conversation based on how interesting they feel? Provide a rating between $0$ to $4$, with $0$ being boring and $4$ being most interesting. This conversation involves a student learning English as a second language from a teacher. Assume the role of the student. Rate as follows: INT: `interest in the teacher's reply,' EXP\_INT: `expected interest in the next conversation,' REASON: `justify the rating.' Consider previous conversations. Next Dialogue: \textit{[Teacher and student dialogue snippet]}
\end{tcolorbox}

The prompt used for instructing off-the-shelf base LLMs to rate the turn-level INT and EXP INT:
\begin{tcolorbox}[colback=gray!10, colframe=black!50, title=Prompt for instructing base LLMs]
    \small The student said \textit{[Teacher and student dialogue snippet]}. To evaluate how interesting the message of the student is, the teacher gave an interestingness score from: 0 = not interesting, 1 = slightly interesting, 2 = interesting, 3 = very interesting, 4 = extremely interesting. The teacher gave:
\end{tcolorbox}

\section{Metric Implementation}
Traditional readability metrics were computed using the \textit{textstat} library\footnote{\href{https://pypi.org/project/textstat/}{https://pypi.org/project/textstat/}} ). The GIS score was computed from the GisPy package \cite{hosseini-etal-2022-gispy}, a Python toolkit for extracting psycholinguistic features. The model-based estimates of conversational uptake are based on Edu-Convokit \cite{wang2024convokit}.

\section{Dataset Information}
The dataset is intended for research purposes only and is released in compliance with the original access conditions set by Prolific and the ethics approval guidelines.
The dataset has been fully anonymized, with all personally identifiable information removed. Participants were informed about anonymization procedures and data protection measures.
The dataset is documented with details on domains, languages, linguistic phenomena, and demographic information, ensuring transparency and reproducibility.

\section{Annotation Interface}
\label{sec:appendix_UI}
\begin{figure*}
    \centering
    \includegraphics[width=0.95\linewidth]{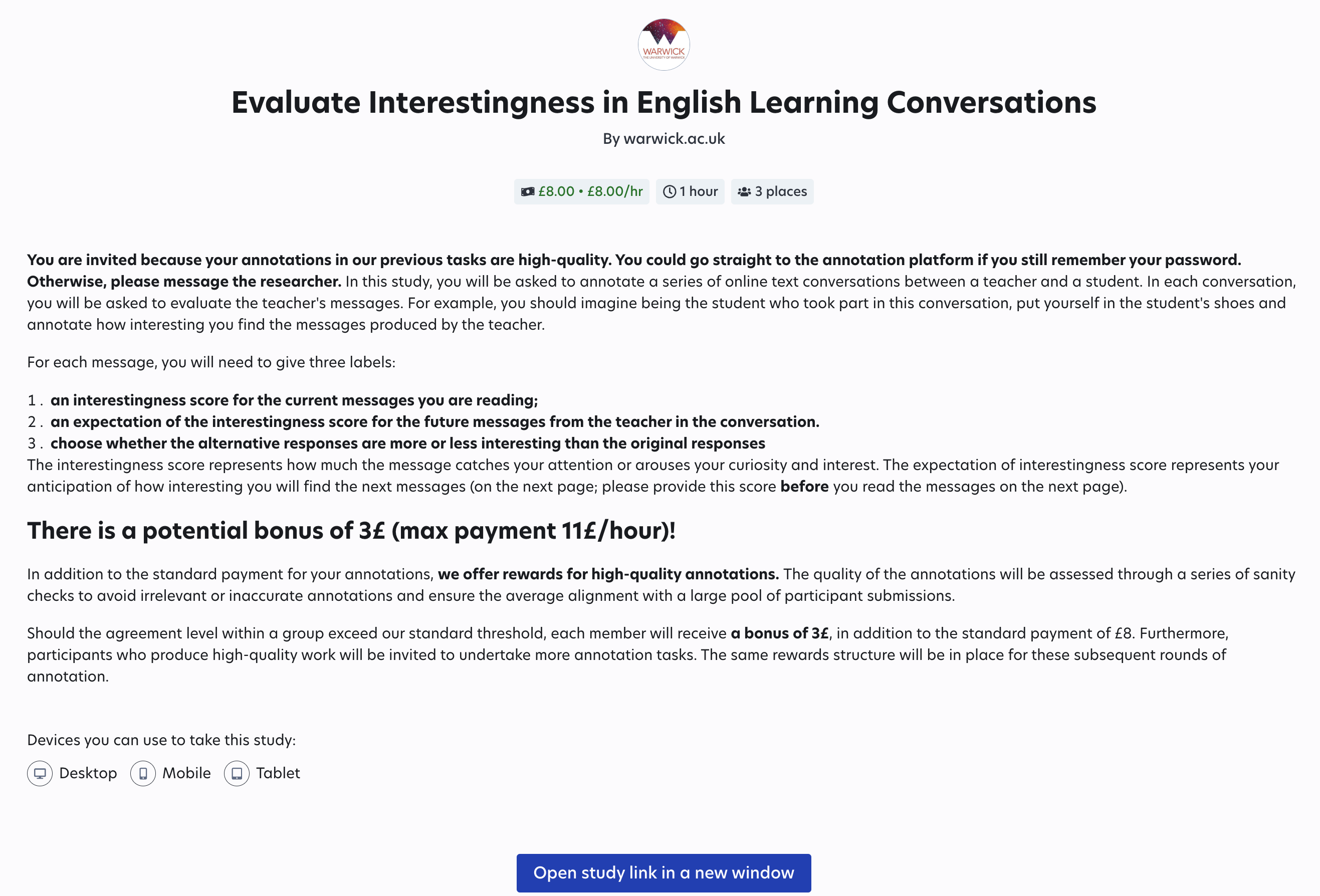}
    \caption{The recruitment page of sequence-level annotation on Prolific.}
    \label{fig:prolific_sequence}
\end{figure*}

\begin{figure*}
    \centering
    \includegraphics[width=0.95\linewidth]{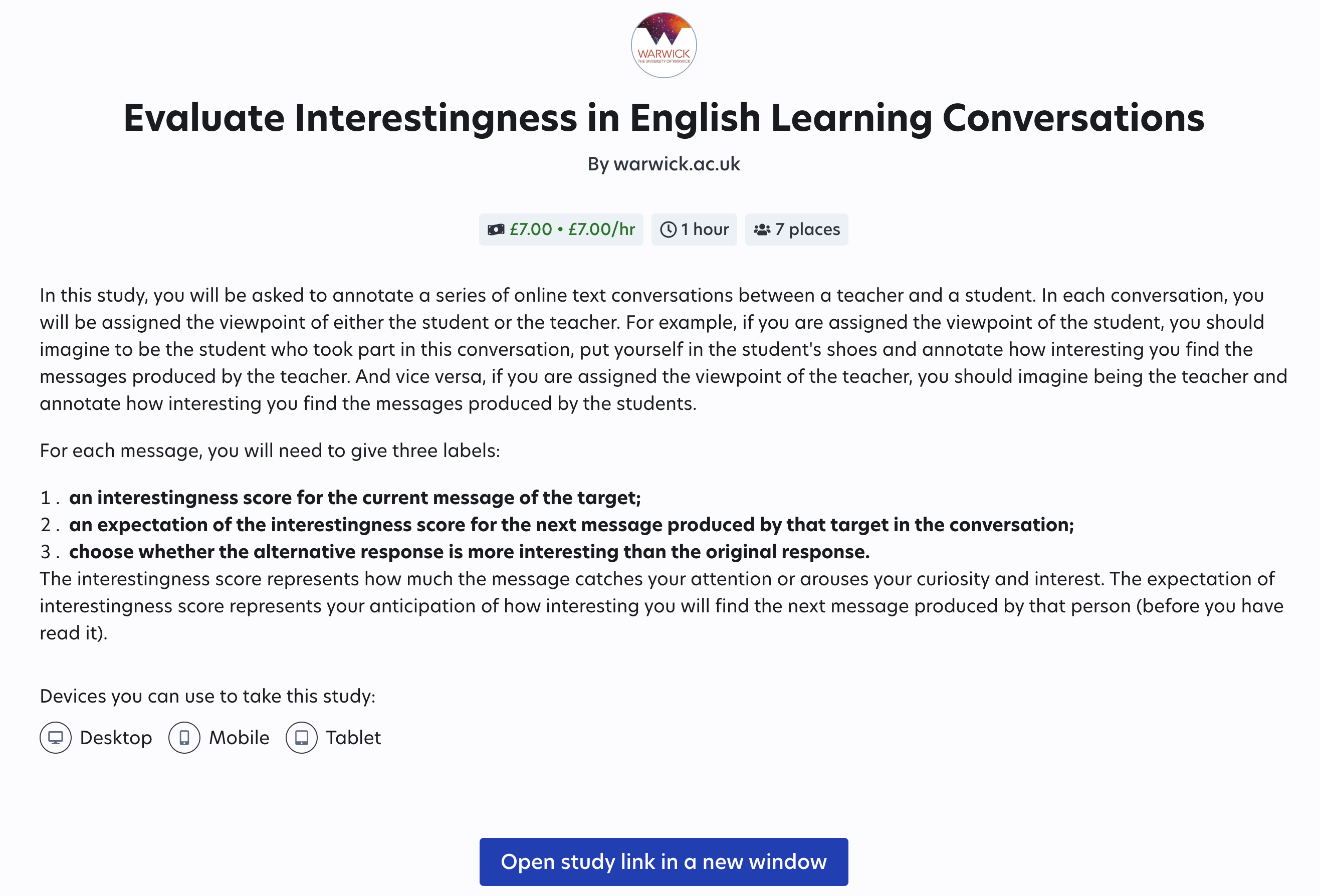}
    \caption{The recruitment page of turn-level annotation on Prolific.}
    \label{fig:prolific_turn}
\end{figure*}

\begin{figure*}[htb]
    \centering
    \includegraphics[width=0.9\linewidth]{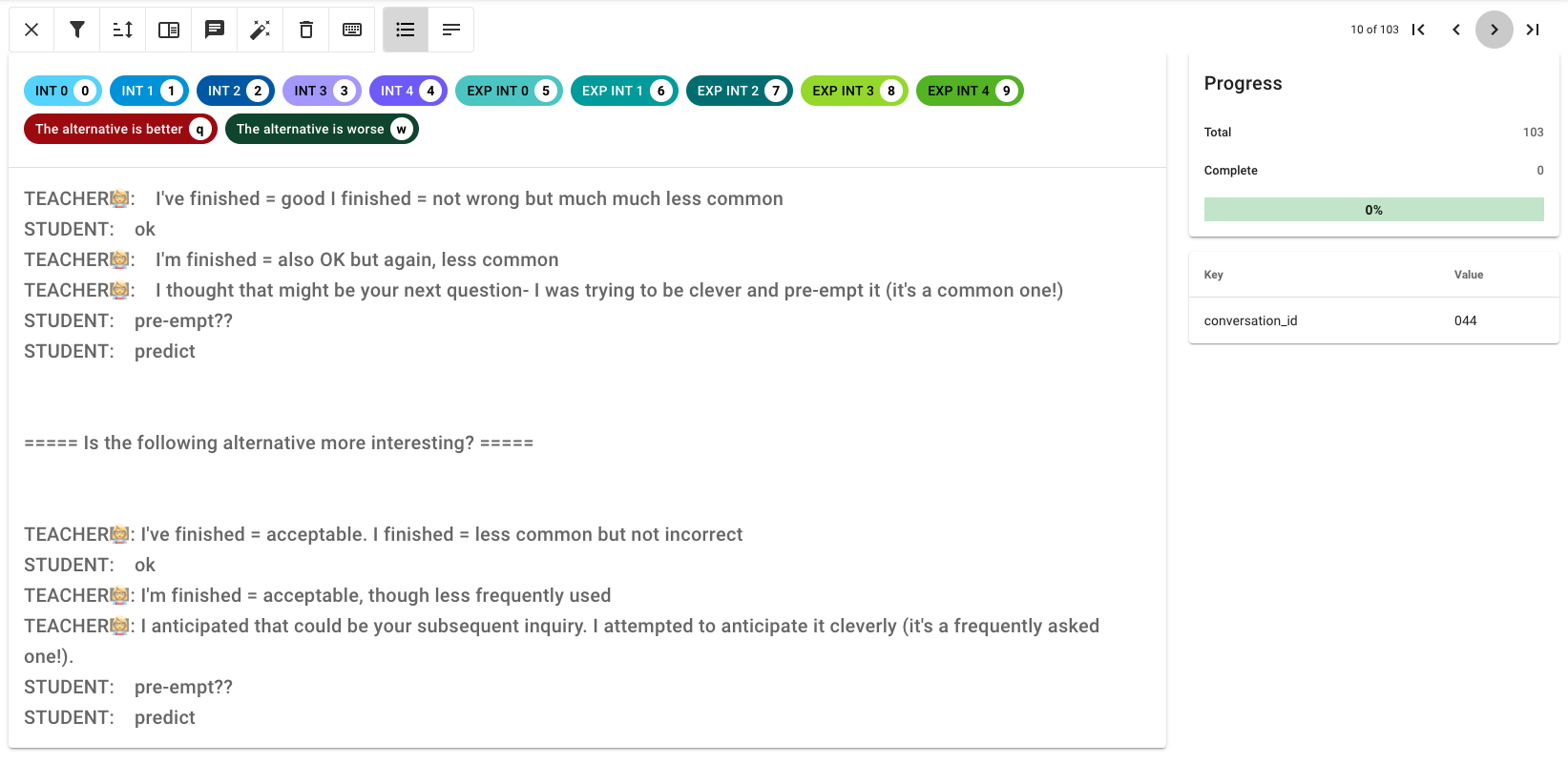}
    \caption{The user interface of a page in the sequence-level annotation task.}
    \label{fig:ui_screenshot}
\end{figure*}

In this section, we provide more details of our recruitment and annotation interface; see Figure \ref{fig:ui_screenshot}.
Figure \ref{fig:prolific_sequence} and \ref{fig:prolific_turn} show the recruitment page on Prolific.

\end{document}